\pdfoutput=1

\documentclass[11pt]{article}

\usepackage{EMNLP2022}

\usepackage{times}
\usepackage{latexsym}
\usepackage{amsmath}
\usepackage{graphicx}
\usepackage{multirow}
\usepackage{multicol}
\usepackage{arydshln}
\usepackage[hang,flushmargin]{footmisc}
\usepackage{algorithm}
\usepackage{algpseudocode}
\usepackage{booktabs}
\usepackage{xurl}

\newcommand{\reffig}[1]{Figure \ref{#1}}
\newcommand{\reftbl}[1]{Table \ref{#1}}
\newcommand{\refsec}[1]{Section \ref{#1}}
\newcommand{\refapp}[1]{Appendix \ref{#1}}
\definecolor{coralpink}{rgb}{0.97, 0.51, 0.47}

\usepackage[T1]{fontenc}

\usepackage[utf8]{inputenc}

\usepackage{microtype}

\usepackage{inconsolata}

\title{Evaluating the Diversity, Equity and Inclusion of NLP Technology:\\ A Case Study for Indian Languages}

\author{Simran Khanuja\thanks{{ } Equal contribution.} \\
  Carnegie Mellon University\thanks{{ } Work done at Google Research.} \\
  \texttt{skhanuja@andrew.cmu.edu} \\\And
  Sebastian Ruder\footnotemark[1] \\
  Google Research \\
  \texttt{ruder@google.com} \\\And
  Partha Talukdar \\
  Google Research \\
  \texttt{partha@google.com} \\}

\begin{document}
\maketitle

\begin{abstract}
In order for NLP technology to be widely applicable, fair, and useful, it needs to serve a \emph{diverse} set of speakers across the world's languages, be \emph{equitable}, i.e., not unduly biased towards any particular language, and be \emph{inclusive} of all users, particularly in low-resource settings where compute constraints are common. In this paper, we propose an evaluation paradigm that assesses NLP technologies across all three dimensions. While diversity and inclusion have received attention in recent literature, equity is currently unexplored. We propose to address this gap using the Gini coefficient, a well-established metric used for estimating societal wealth inequality. Using our paradigm, we highlight the distressed state of current technologies for Indian (IN) languages (a linguistically large and diverse set, with a varied speaker population), across all three dimensions. To improve upon these metrics, we demonstrate the importance of region-specific choices in model building and dataset creation, and more importantly, propose a novel, generalisable approach to optimal resource allocation during fine-tuning. Finally, we discuss steps to mitigate these biases and encourage the community to employ multi-faceted evaluation when building linguistically diverse and equitable technologies.
\end{abstract}

\section{Introduction}
\label{intro}


NLP has seen large advances in recent years driven by the rapid progress in transfer learning \cite{ruder-etal-2019-transfer,Devlin2019}. The benefits of these advances, however, are not equally distributed across the world’s languages \cite{Joshi2020} and users. While linguistic diversity and inclusion have evolved to be a pressing concern today, measures to quantify these are still lacking. The progress of any field is tightly coupled with its evaluation paradigm and the community is incentivized to work on highly visible metrics and benchmarks. In order for users around the world to reap the benefits of NLP technology, we must move from an evaluation that focuses on optimizing raw performance on available test data to a more holistic user-centric evaluation \cite{Ethayarajh2020,Ruder2021}. In this paper, we attempt to do so by defining an evaluation framework along three dimensions: diversity, equity, and inclusion.\footnote{We focus on assessing these dimensions on the \emph{language level}. Prior work on equity focuses mainly on subpopulations \emph{within} a language \cite{Katell2020}.}

Diversity is important as NLP technology should be available to speakers of any language \cite{elra}. To this end, recent work \cite{blasi2021systematic} quantifies diversity of NLP technology across the world's languages by weighing normalized task performance for each language based on its speaker population. 


Equity is key as we should aim to develop technology that does not discriminate against speakers of any particular language \cite{kaneko-bollegala-2019-gender}. State-of-the-art multilingual models in fact have been shown to perform much better in languages with access to many pre-training resources \cite{Hu2020xtreme}. To measure such performance inequity across languages, we propose to use the Gini coefficient \cite{dorfman1979formula}, a measure that has been used to represent the income inequality within social groups.

Finally, inclusion is a concern as the fact that NLP technology is performant in a given task and language does not mean that it is usable by all. State-of-the-art models are becoming larger and larger \cite{Fedus2021} and the low-resource settings of many languages often coincide with constraints on computational resources \cite{Ahia2021}. The value a technology provides to a user thus also needs to consider how easily such technology can be deployed in practice. \citet{ma2021dynaboard} quantify this based on a model's runtime efficiency, considering factors like throughput and memory.

Our proposed paradigm is language and model agnostic making it applicable to an arbitrary set of languages and models. We apply our paradigm to highlight the distressing state of current technologies for Indian (IN) languages. India is a multilingual society with 1369 rationalized languages and dialects being spoken across the country \cite{india2011census}. Of these, 22 scheduled languages\footnote{Assamese, Bengali, Bodo, Dogri, Gujarati, Hindi, Kashmiri, Kannada, Konkani, Maithili, Malayalam, Manipuri, Marathi, Nepali, Oriya, Punjabi, Tamil, Telugu, Sanskrit, Santali, Sindhi, Urdu} spoken by almost 97\% of the population hold an official recognition and 121 languages have more than 10,000 speakers. Additionally, 21.92\% of its population lives below the poverty line \cite{rbi-stats-poverty}. Serving this large varied population justly requires a multi-faceted effort and basing our case study on IN languages directs the way forward. 

We evaluate a range of state-of-the-art models and transfer settings \cite{Hu2020xtreme} across four standard downstream tasks: \emph{Named Entity Recognition} (NER), \emph{Part-of-Speech Tagging} (POS), \emph{Natural Language Inference} (NLI) and \emph{Question Answering} (QA). We observe that region-specific choices, i.e., a) region-specific pre-trained models \cite{kakwani2020inlpsuite,khanuja2021muril} and b) Hindi as the transfer language during fine-tuning, generally yield the best results. In terms of efficiency, we find that smaller models are preferable for easier, syntactic tasks while larger models have the edge on more complex, semantic tasks.

Our findings, however, also highlight that we are still a long way from building perfectly inclusive and equitable NLP technology. Towards bridging this gap, we explore how we can most effectively annotate data for the remaining languages. Past work \cite{lin2019choosing,ahuja-etal-2022-multi} has relied on heuristic and feature-based approaches to source language selection. In our work, we propose a novel, fully computational approach to model the space of source and target languages, and derive the optimal allocation of a fixed annotation budget to maximize performance on our proposed metrics in a multi-source setting.

Our contributions are the following: \textbf{1)} We propose a holistic evaluation paradigm that assesses NLP technology based on their diversity, equity, and inclusion. \textbf{2)} Using this paradigm, we evaluate model capabilities for IN languages and quantify their shortcomings. \textbf{3)} We propose a novel approach to select data for fine-tuning these models with the objective of maximizing performance on the proposed metrics. \textbf{4)} We discuss steps that must be taken to mitigate these biases and call upon the community to incorporate our evaluation paradigm when building models to track progress towards building linguistically inclusive and diverse technologies.

\section{Background and Related Work}
\paragraph{Multilingual Models} Transformer-based language models (LMs) \cite{vaswani2017attention} trained on massive amounts of text from multiple languages have enabled the inclusion of an unprecedented number of languages in NLP technologies \cite{conneau2019unsupervised, devlin2018bert}. However, previous research has shown that these models do not serve all languages equally, with resource-poor languages in the long tail suffering the most \cite{Hu2020xtreme, lauscher2020zero}. These models go through a critical step of fine-tuning for the downstream task before being deployed. Several recent works focus on optimal fine-tuning strategies that mitigate transfer gaps and improve overall performance across target languages. \citet{lin2019choosing} propose a tool that chooses optimal transfer languages based on linguistic features. \citet{lauscher2020zero} demonstrate the effectiveness of investing in few-shot in-language training examples. Recently, \citet{debnath2021towards} show that investing in an equal number of fine-tuning instances across target languages performs best. These past approaches however, have all been heuristically designed based on the knowledge and intuition of the experimenter.

\paragraph{User-centric Evaluation} At its core, the need for language diversity in technologies is tied to the people it serves. Previous work \cite{Ethayarajh2020,ma2021dynaboard} has highlighted the need for transparent and user-centric leaderboard evaluation, reporting practically relevant statistics such as model size, energy efficiency, and inference latency. It is common for speaker populations of under-represented languages to operate in resource-constrained settings. Therefore, in addition to evaluating \emph{linguistic} diversity, we follow \citet{ma2021dynaboard} in computing model efficiency, which serves to assess the \emph{inclusivity} of these technologies. With regards to linguistic diversity, \citet{Ruder2021} highlight the need for more fine-grained evaluation across languages and introduce language-specific leaderboards. \citet{blasi2021systematic} quantify the value of NLP technology weighed by speaker population and determine utilities of several technologies across the world's languages. \citet{choudhury2021linguistically} propose strategies for fair and efficient model selection depending on one's application, based
on the principles of fairness in economics and social choice theory.


\paragraph{Indian Languages} The research community has actively been contributing to the advancement of IN NLP by collecting and open-sourcing data \cite{kakwani2020inlpsuite, ramesh2021samanantar, abraham2020crowdsourcing, roark-etal-2020-processing, kunchukuttan2017iit, khanuja2020new}, building region-specific multilingual models \cite{khanuja2021muril, kakwani2020inlpsuite, ramesh2021samanantar} and creating evaluation benchmarks \cite{kakwani2020inlpsuite, khanuja2020gluecos}\footnote{\footnotesize \href{https://github.com/AI4Bharat/indicnlp\_catalog}.{https://github.com/AI4Bharat/indicnlp\_catalog} maintains a list of resources for Indian NLP.} Several of these efforts have been undertaken by AI4Bharat\footnote{\href{https://ai4bharat.org/}{https://ai4bharat.org/}}, a non-profit open-source community that has additionally been working on developing resources for IN signed languages \cite{sridhar2020include} and creating keyboards for IN scripts. Recently, Google Research India launched a question answering (QA) challenge named ChAII\footnote{\href{https://www.kaggle.com/c/chaii-hindi-and-tamil-question-answering}{\footnotesize https://www.kaggle.com/c/chaii-hindi-and-tamil-question-answering}}. 
Microsoft Research India has also made significant contributions to IN NLP with several efforts directed towards code-mixed language processing\footnote{\href{https://www.microsoft.com/en-us/research/project/melange/}{https://www.microsoft.com/en-us/research/project/melange}} and building tools and datasets for under-represented languages in India\footnote{\href{https://www.microsoft.com/en-us/research/project/ellora/}{https://www.microsoft.com/en-us/research/project/ellora}}.

\section{Diversity, Equity and Inclusion (DEI)}
There is increasing awareness in society to promote diversity, equity and inclusion in our workforce, wherein such measures have recently been enforced by law \cite{constitution}. In the social construct, \emph{diversity} is defined as ``the practice of including the many communities, identities, races, ethnicities, backgrounds, abilities, cultures, and beliefs of the people, including underserved communities'', \emph{equity} refers to ``the consistent and systematic fair, just, and impartial treatment of all individuals'' and \emph{inclusion} means ``the recognition, and use of the talents and skills of employees of all backgrounds'' \cite{constitution}. Given the ubiquitous use of technology in our daily lives, we as technology makers hold the responsibility of making sure all voices are heard and equally represented in the technology we serve. Given that our research community is incentivized to work on highly visible metrics and benchmarks, an important first step is to encourage evaluation along these dimensions. Previous work mainly focused on average performance (as measured by accuracy or F1 for NLU tasks), which is not indicative of differences in DEI. Hence, while models claim state of the art based on an increase in average performance, this increase may only be due to making the ``rich richer'' (see \reftbl{table:baseline}).

We propose an evaluation paradigm for current NLP technology that operationalizes the well-established diversity, equity and inclusion pillars on a language level: we quantify diversity based on the value diverse speaker populations derive from a technology, equity based on egalitarian performance across speaker populations, and inclusion based on a technology's accessibility. We employ metrics of \citet{blasi2021systematic} and \citet{ma2021dynaboard} to measure diversity and inclusion respectively and propose a new metric to quantify equity. We describe the metrics in more detail below:

\label{definitions}
\subsection{Diversity: \emph{Utility, Demand and the Global Metric}}
\label{global_metric_def}
The global metric introduced by \citet{blasi2021systematic} helps quantify linguistic diversity. Formally, this metric is composed of the utility of a technology weighed by its demand. The utility $\mathrm{u_l}$ of a system for a task and language is its performance normalized by the best possible performance (typically, human-level performance) afforded by the task:
\begin{align*}
\mathrm{u_l = \frac{\text{performance}_l}{\text{theoretical max performance}}}
\label{utility_eqn}
\end{align*}
%

Demand $\mathrm{d_l}$ is characterized by taking into consideration demographic and linguistic perspectives. Under the demographic perspective, the demand for a given technology in a language is estimated to be proportional to the number of speakers of the language itself $\mathrm{n_l~(d_l \propto n_l)}$. Under the linguistic  perspective, the demand across languages is identical $\mathrm{(d_l \propto 1)}$. These two alternatives,  as well as any intermediate combination of them, are parameterized through a single exponent $\mathrm{\tau}$:
\begin{align*}
\mathrm{d_l^{(\tau)} = \frac{n_l^{\tau}}{\sum_{l' \epsilon L}\ n_{l'}^{\tau}}}
\end{align*}
\noindent where $\mathrm{\tau = 1}$ corresponds to a demographic notion of demand and $\mathrm{\tau = 0}$ to a linguistic one. The global metric can now be defined as:
\begin{align*}
\mathrm{M_{\tau} = \sum_{l \epsilon L}\ d_l^{(\tau)} .\ u_l}
\end{align*}

\noindent In essence, $\mathrm{M_{\tau}=0}$ means that no user benefits from language technology and $\mathrm{M_{\tau}=1}$ corresponds to each language user enjoying perfect technology.
Given our people-centric aim to measure benefit for all speakers, we employ the demographic notion of demand ($\mathrm{M_{\tau=1}}$).

\begin{figure}
\includegraphics[scale=0.175]{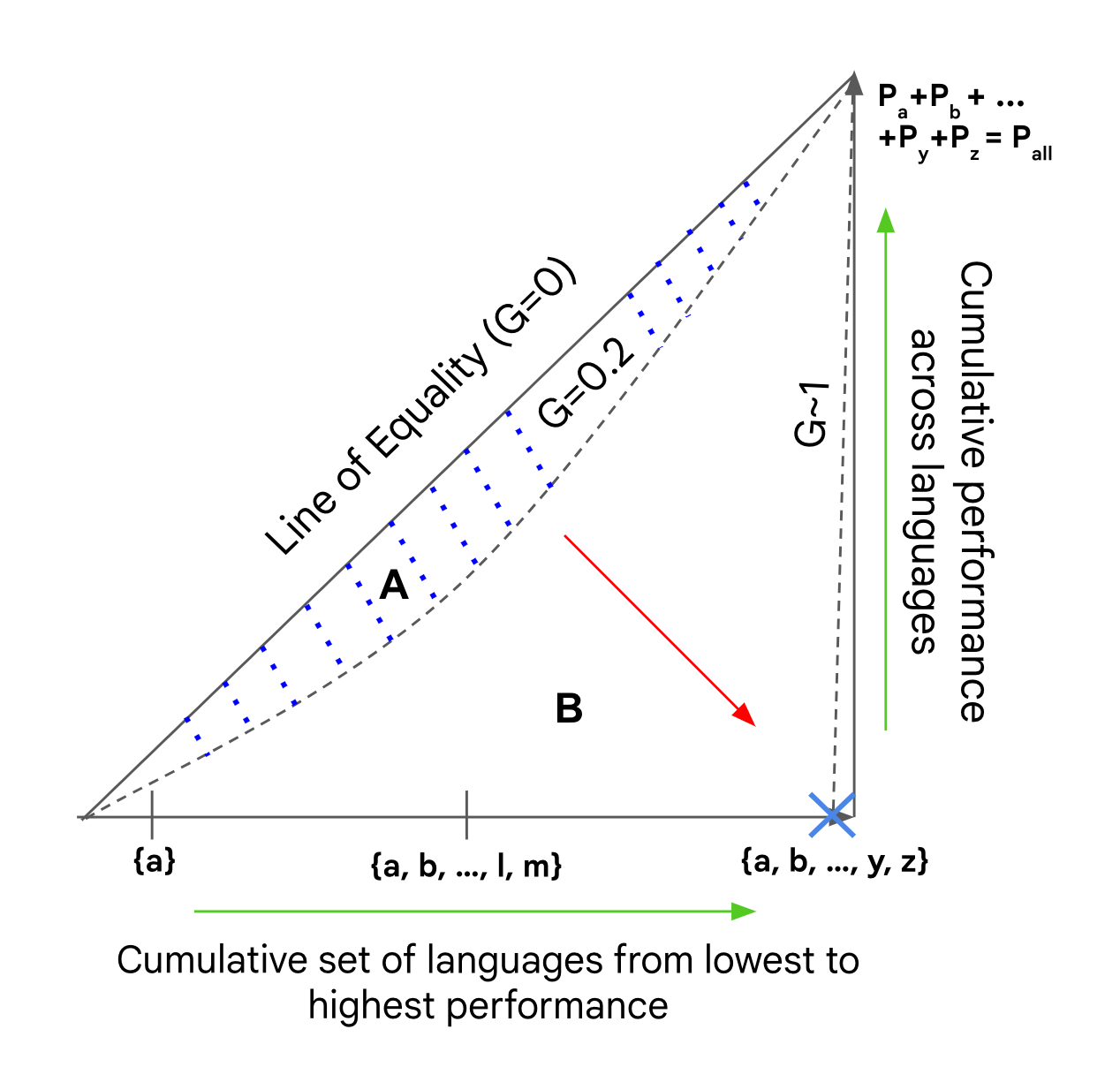} 
\caption{\emph{Graphical Representation} of the Gini coefficient ($\mathrm{G}$), given by $\mathrm{A / (A+B)}$ when $\mathrm{G=0.2}$. Each point on the graph depicts the proportion of the total (cumulative) performance $P_{all}$ (e.g., accuracy, F1, etc) that is achieved by the bottom $n\%$ of languages combined. Assume we have a language set \{a, b, \ldots,  y, z\} with performances $\mathrm{P_{a} \leq P_{b} \: ... \leq P_{y} \leq P_{z}}$ and $\mathrm{P_{a} + P_{b} \: ...  \: P_{y} + P_{z} = P_{all}}$. When all languages perform the same, i.e., $\mathrm{P_{a}=P_{b}= ... =P_{y}=P_{z}}$, $\mathrm{G=0}$, as represented by the line of equality, i.e., the bottom $n$\% of languages also account for $n$\% of the total performance. The value of $\mathrm{G}$ increases as the disparity in performance between all languages increases, and approaches unity in the case of perfect inequality (here, this would mean $\mathrm{P_{a}=P_{b}= ... =P_{y}=0}$ and $\mathrm{P_{z}=P_{all}}$), i.e., the model / application supports only one language. See \textsection \ref{sec:gini} for details.}
\label{fig:gini}
\end{figure}

\subsection{Equity: \emph{Gini Coefficient}}
\label{sec:gini}

While diversity accounts for a language's speaker population, it does not take into account inequalities in the performance across languages. While several past works have highlighted transfer gaps in performance across languages \cite{Hu2020xtreme}, none have quantified this dispersion.\footnote{\citet{Hu2020xtreme} only considered the difference between English and other languages as cross-lingual transfer gap.} Traditionally used measures of statistical dispersion like standard deviation or calculating range are sub-optimal choices as they are scale-dependant, unbounded and highly sensitive to outliers, which makes them unsuitable for data that does not approach a normal distribution \cite{de2007income}. 

Beyond these measures, several nuanced metrics have been introduced to quantify disparity in income distributions. The choice of income inequality indicator is of significant importance since it has implications in measuring health, state-level mortality, etc. \cite{de2007income}. The Gini coefficient \cite{dorfman1979formula} has been most commonly used for this purpose \cite{de2007income}.

\citet{hurley2009comparing} lay out six desirable attributes of a measure of sparsity, drawing from past literature \cite{dalton1920measurement, rickard2004gini} and prove the Gini coefficient to be the only measure having all six, among a varied set of alternatives. Briefly, these properties and their relevance in measuring linguistic disparity across tasks include: i) \emph{Robin Hood}: a drop in high-performant and gain in low-performant languages should lead to higher equity; ii) \emph{Scale Invariance}: no change in relative performance should lead to no change in equity, regardless of changes in absolute values; iii) \emph{Rising Tide}: adding a constant value to each language's performance should increase equity; iv) \emph{Cloning}: equity must remain invariant under cloning, i.e., if two identical distributions are combined, the equity remains unchanged; v) \emph{Bill Gates}: if one language hypothetically gains infinite performance, equity should tend to zero; vi) \emph{Babies}: adding languages with zero performance in the distribution should decrease equity.


Given that the Gini coefficient satisfies all of these attributes \cite{hurley2009comparing}, we propose to use the same in pursuit of quantifying the inequalities amongst languages with regard to downstream tasks in NLP. A pictorial representation of the Gini coefficient for this setting can be found in \reffig{fig:gini}. Downstream task performance closely follows the highly skewed data distributions on which massively multilingual models are pre-trained. By including the Gini coefficient measure in our evaluation, we aim to incentivize model builders to invest in equitable performance, despite data differences. Details on how the Gini coefficient is calculated are given in \refapp{app:gini}.

\begin{table*}[ht]
\centering
\scalebox{0.9}{
\begin{tabular}{c c c c c c c c c c c c c}
\toprule
\textbf{Language} & as & bn & brx & doi & en & gu & hi & kn & kok & ks & mai & ml \\
\textbf{Speakers (in M)} & 23.6 & 107.4 & 1.6 & 2.8 & 128.5 & 60.3 & 691.6 & 58.8 & 2.6 & 7 & 14.3 & 35.6 \\ \midrule
\textbf{Language} & mni & mr & ne & or & pa & sa & sat & sd & ta & te & ur & - \\
\textbf{Speakers (in M)} & 2.2 & 99.1 & 3.4 & 42.6 & 36.1 & 3.1 & 7.7 & 3.1 & 76.6 & 94.5 & 63.2 & - \\
\bottomrule
\end{tabular}}
\caption{The number of speakers (in millions) for each of the 22 scheduled languages and English. We take the sum total of first, second and third language speakers for each language.}
\label{speakers}
\end{table*}

\subsection{Inclusion: \emph{Efficiency Score}}
\label{sec:efficiency}

Language technology is only beneficial if it can be deployed and accessed by users in a region. We employ efficiency to quantify inclusion as user devices are resource-constrained in many low-resource settings. Following work on user-centric evaluation \cite{Ethayarajh2020,ma2021dynaboard}, we propose to incorporate efficiency into model performance based on throughput and memory, each of which are defined below.\footnote{\citet{ma2021dynaboard} additionally consider fairness and robustness, both of which are highly contextual and difficult to define in the context of multilingual models at present. Hence, we focus on model aspects that are objectively measurable.}

\paragraph{Throughput} Number of instances the model can process per second on a CPU, assuming that GPUs are rarely used for deployment at scale in resource-constrained environments.


\paragraph{Memory Saved} The size of the model is considered to be a measure of how expensive a model is to use in practice. Since we wish to minimize this metric, \emph{memory used} is transformed into \emph{memory saved} by subtracting it from a maximum available memory of 16 GB \cite{ma2021dynaboard}. We show the memory and throughput values for our models in \refapp{appendix:efficiency}. 

Following \citet{ma2021dynaboard}, to calculate the efficiency score we first convert each metric into units of performance, by calculating the average marginal rate of substitution ($\mathrm{AMRS}$) for each metric $\mathrm{M}$ (i.e., throughput and memory). $\mathrm{AMRS (M, perf)}$ tells us the rate at which model creators, as a group, are trading off $\mathrm{M}$ for a one-point increase in $\mathrm{perf}$ while keeping utility constant. For example, if $\mathrm{AMRS}$ of “memory saved” with respect to accuracy
were 0.5 GB, then each GB of memory saved would on average be worth 2 points of accuracy. Dividing $\mathrm{M}$ by $\mathrm{AMRS(M, perf)}$ converts it to units of performance. Details on how one can calculate $\mathrm{AMRS(M, perf)}$ can be found in \refapp{appendix:efficiency}. For a model $\mathrm{x_i}$, $\mathrm{Efficiency(x_i)}$ is then defined as :
\begin{align*}
& \mathrm {Efficiency(x_i) = \sum_{M} w_{M} * \frac{M(x_i)}{AMRS(M, perf)}} \end{align*}

\noindent where we choose $\mathrm{w_{perf}}=0.5$, $\mathrm{w_{throughput}}=0.25$ and $\mathrm{w_{memory}}=0.25$ as default weights. In practice, these weights can be adjusted based on user requirements and existing constraints.

\section{Egalitarian Annotation Budget Allocation} \label{sec:greedy}

Model development involves not just the design of an architecture or training but also data annotation. The proposed dimensions thus cannot only be used to assess models but can also inform how data should be annotated across many languages. As fine-tuning on a few labeled examples in the target language has shown to improve zero-shot transfer performance, we study how to allocate an annotation budget across a number of source languages $\mathrm{S}$ in order to optimize for inclusion and equity across a set of target languages $\mathrm{T}$. Previous work employs a feature-based approach to select a single source language to maximize performance on a target language \cite{lin2019choosing} or labels examples across all source languages equally \cite{debnath2021towards}. We propose a fully computational approach for modeling the space of source and target languages for a multi-source multi-target language setting. This is done by empirically estimating performance of language $\mathrm{t\ \epsilon\ T}$ on a held-out set when fine-tuned on $\mathrm{x}$ labeled instances of language $\mathrm{s\ \epsilon\ S}$, $\mathrm{\forall (s, t)}$ pairs, which follows a power-law distribution \cite{rosenfeld2019constructive}. 
We now seek to find the optimal allocation 
$\mathrm{\{x_s : s \in S\}}$ subject to $\mathrm{\sum_{s \in S} x_s \le X}$ (details in \refapp{sec:budget_alloc}).

We follow a simple greedy approach to solve this constrained optimization problem as shown in \reftbl{algorithm_1}. Specifically, at each step we allocate a sample to the source language conferring the highest marginal gain to all target languages, which is quantified by the summation of the increase in the global metric and the reduction in Gini.\footnote{Future work may consider more complex approaches that consider language relatedness based on work on transfer relationship learning \cite{Zamir2018,Song2019}.} At present, we assign equal weight to each metric but this can be changed according to user preferences.

\section{Experiments}
\label{sec:experiments}

\subsection{Experimental setup}
\label{sec:setup}

\paragraph{Languages} We base our case study on the 22 scheduled languages of India spoken by 97\% of its population. We also include English, since it has a sizeable population of 128.5M speakers (Table \ref{speakers}). 

\paragraph{Tasks} We select tasks from the XTREME \cite{Hu2020xtreme} benchmark. Dataset details and the human performance (HP) for each task can be found in \reftbl{datasets}. For each task, we only evaluate on IN language test sets.

\begin{table}[t]
\centering
\scalebox{0.75}{
\begin{tabular}{p{0.1\linewidth} p{0.45\linewidth} p{0.25\linewidth} p{0.1\linewidth}}
\toprule
\centering Task & \centering Dataset & \centering Test Langs. & HP \\ \midrule
\centering NER & WikiAnn \cite{pan2017cross, rahimi2019massively} & bn, en, gu, hi, ml, mr, pa, ta, te, ur & 97.6 \\
\centering POS & Universal Dependencies v2.6 \cite{nivre:hal-01930733} & en, hi, mr, ta, te, ur & 97 \\
\centering NLI & XNLI  \cite{conneau2018xnli} & en, hi, ur & 92.8 \\
\centering QA & XQuAD \cite{artetxe2019cross}; TyDiQA-GoldP \cite{clark2020tydi} & bn, en, hi, te & 91.2; 90.1  \\
\bottomrule
\end{tabular}}
\caption{\label{datasets} \emph{Finetuning Tasks and Datasets}. HP denotes the human performance for each task. For QA, HP is 91.2 F1 for XQuAD and 90.1 F1 for TyDiQA.}
\end{table}

\paragraph{Models} Model selection is motivated by two key factors that we wish to explore in our study:  \textbf{i)} general vs region-specific choices; and \textbf{ii)} model efficiency. We choose IndicBERT \cite{kakwani2020inlpsuite}, MuRIL \cite{khanuja2021muril} and XLM-R \cite{conneau2019unsupervised}, the first two being region-specific models and the third being a state-of-the-art model trained on 100+ languages. We consider both the base and large versions for MuRIL and XLM-R. IndicBERT follows the ALBERT architecture \cite{lan2019albert} and is hence much smaller than the base versions of both models. IndicBERT is trained on 11, MuRIL on 16, and XLM-R on 15 IN languages (details in \refapp{app:pretraining}).

\begin{table}[t]
\centering
\scalebox{0.7}{
\begin{tabular}{c c c c c}
\toprule
Language & NER & POS & NLI & QA \\ \midrule
English & 20,000 & 21,261 & 392,702 & 88,602 \\
Hindi & 5,000 & 13,305 & 392,702 (\emph{-tran}) & 88,602 (\emph{-tran}) \\
\bottomrule
\end{tabular}}
\caption{\label{train_instances} \emph{Number of training instances for English and Hindi.} (\emph{-tran}) denotes that the English fine-tuning set has been translated to Hindi.}
\end{table}

\begin{table}[t]
\centering
\scalebox{0.55}{
\begin{tabular}{c c c c c c c}
\toprule
\multirow{2}{*}{Task} & \multirow{2}{*}{Model} & Baseline & Diversity & Equity & Inclusion \\ 
& & \emph{F1/Accuracy $\boldsymbol{\uparrow}$} & \emph{M$_{\tau = 1}$ $\boldsymbol{\uparrow}$} &  \emph{Gini Coeff. $\boldsymbol{\downarrow}$} & \emph{Efficiency $\boldsymbol{\uparrow}$} \\ \midrule
\multirow{3}{*}{NER} & $\mathrm{MuRIL_{base}}$ & 77.6 & 69.6 & 0.59 & 69.1 \\
& $\mathrm{XLM}$-$\mathrm{R_{large}}$ & 68.0 & 61.2 & 0.60 & 44.4 \\
& $\mathrm{MuRIL_{large}}$ & 77.7 & 68.2 & 0.59 & 63.1 \\
\midrule
\multirow{3}{*}{POS} & $\mathrm{MuRIL_{base}}$ & 75.0 & 54.7 & 0.76 & 52.5 \\
& $\mathrm{XLM}$-$\mathrm{R_{large}}$ & 79.2 & 60.3 & 0.75 & 48.0 \\
& $\mathrm{MuRIL_{large}}$ & 77.3 & 58.6 & 0.76 & 51.8 \\
\midrule
\multirow{3}{*}{NLI} & $\mathrm{MuRIL_{base}}$ & 74.1 & 45.5 & 0.88 & 58.7 \\ 
& $\mathrm{XLM}$-$\mathrm{R_{large}}$ & 78.7 & 46.6 & 0.88 & 57.3 \\
& $\mathrm{MuRIL_{large}}$ & 78.6 & 47.4 & 0.88 & 57.8 \\
\midrule
\multirow{3}{*}{QA} & $\mathrm{MuRIL_{base}}$ & 76.1 & 53.8 & 0.83 & 77.8 \\
& $\mathrm{XLM}$-$\mathrm{R_{large}}$ & 75.7 & 56.6 & 0.83 & 76.3 \\ 
& $\mathrm{MuRIL_{large}}$ & 77.7 & 57.9 & 0.83 & 75.7 \\
\bottomrule
\end{tabular}}
\caption{\label{table:baseline}\emph{DEI Results compared to baseline F1/accuracy performance.} Here, we compare models' accuracy/F1 performances (usually reported as the evaluation metric) to their DEI metrics. We observe that while performances may significantly vary, DEI metrics (especially equity) don't change as much, indicating that multilingual models make the rich "richer" to increase average performance but may not be moving towards being truly multilingual and equitable across languages. More discussions in \refsec{sec:zero_shot}.}
\end{table}

\begin{table}[t]
\centering
\scalebox{0.55}{
\begin{tabular}{c c c c c c c c}
\toprule
Metric & Train Lang. & Model & NER & POS & NLI & QA & Average \\ \midrule
\multirow[b]{3}{*}{$\mathrm{M_{\tau = 1} \boldsymbol{\uparrow}}$} & \multirow{3}{*}{English}& $\mathrm{MuRIL_{base}}$ & 69.6 & 54.7 & 45.5 & 53.8 & 55.9 \\
& & $\mathrm{XLM}$-$\mathrm{R_{large}}$ & 61.2 & 60.3 & 46.6 & 56.6 &  56.2 \\
& & $\mathrm{MuRIL_{large}}$ & 68.2 & 58.6 & 47.4 & \textbf{57.9} & 58.0 \\
\cline{2-8} \multirow[t]{3}{*}{(Diversity)} & \multirow{3}{*}{Hindi} & $\mathrm{MuRIL_{base}}$ & \textbf{75.1} & \textbf{67.3} & 46.8 & 54.7 & 61.0 \\
& & $\mathrm{XLM}$-$\mathrm{R_{large}}$ & 74.4 & 66.8 & \textbf{49.4} & 53.2 & 60.9 \\
& & $\mathrm{MuRIL_{large}}$ & 74.8 & 66.5 & 49.2 & 54.6 & \textbf{61.3} \\
\midrule
\multirow[b]{3}{*}{Gini Coeff. $\boldsymbol{\downarrow}$} & \multirow{3}{*}{English} & $\mathrm{MuRIL_{base}}$ & 0.59 & 0.76 & 0.88 & 0.83 & \textbf{0.76} \\
& & $\mathrm{XLM}$-$\mathrm{R_{large}}$ & 0.6 & 0.75 & 0.88 & 0.83 & 0.77 \\
& & $\mathrm{MuRIL_{large}}$ & 0.59 & 0.76 & 0.88 & 0.83 & 0.77 \\
\cline{2-8} \multirow[t]{3}{*}{(Equity)} & \multirow{3}{*}{Hindi} & $\mathrm{MuRIL_{base}}$ & 0.59 & 0.75 & 0.87 & 0.83 & \textbf{0.76} \\
& & $\mathrm{XLM}$-$\mathrm{R_{large}}$ & 0.59 & 0.76 & 0.88 & 0.83 & 0.77 \\
& & $\mathrm{MuRIL_{large}}$ & 0.59 & 0.75 & 0.87 & 0.83 & \textbf{0.76} \\
\midrule
\multirow[b]{3}{*}{Efficiency $\boldsymbol{\uparrow}$} & \multirow{3}{*}{English} & $\mathrm{MuRIL_{base}}$ & 69.1 & 52.5 & 58.7 & \textbf{77.8} & 64.5 \\
& & $\mathrm{XLM}$-$\mathrm{R_{large}}$ & 44.4 & 48 & 57.3 & 76.3 & 56.5 \\ 
& & $\mathrm{MuRIL_{large}}$ & 63.1 & 51.8 & 57.8 & 75.7 & 62.1 \\
\cline{2-8} \multirow[t]{3}{*}{(Inclusion)} & \multirow{3}{*}{Hindi} & $\mathrm{MuRIL_{base}}$ & \textbf{69.8} & \textbf{56.2} & \textbf{59.8} & 77.3 & \textbf{65.8} \\
& & $\mathrm{XLM}$-$\mathrm{R_{large}}$ & 49.2 & 49.0 & 59.1 & 75.1 & 58.1 \\
& & $\mathrm{MuRIL_{large}}$ & 65.2 & 53.7 & 58.8 & 75.0 & 63.2 \\
\bottomrule
\end{tabular}}
\caption{\label{table:zero_shot_cut}\emph{Region-specific fine-tuning results.} Note that the metrics are computed considering all 23 languages as detailed in \refsec{sec:setup}. Region-specific fine-tuning helps, but disparities along DEI axes persist. More discussions in \refsec{sec:zero_shot}.}
\end{table}

\paragraph{Fine-tuning} We initially fine-tune the selected models using training data in English (EN) given the availability of labeled data across tasks. However, past works highlight that this choice is sub-optimal and one can obtain better performance by transferring from closely related languages \cite{lauscher2020zero, cotterell2017cross, dong2015multi, turc2021revisiting}. To examine this effect in our case study, we additionally fine-tune models on Hindi (HI) because \textbf{i)} 15 out of 22 languages belong to the same language family as HI (Indo-Aryan); \textbf{ii)} we have training data available for all tasks in HI\footnote{Training sets for NLI and QA have been machine-translated from English, which has been shown to perform similar to human-generated train sets \cite{turc2021revisiting}.}; and \textbf{iii)} HI has the highest speaker population, which may lead to higher demographic utility and is a future-safe choice to obtain annotations for any task. \reftbl{train_instances} summarizes training data statistics for EN and HI.


\subsection{Zero-shot transfer results}
\label{sec:zero_shot}

\noindent \textbf{How do DEI metrics compare to baseline standard performance metrics (F1/Accuracy)?} We report results of the best-performing models in \reftbl{table:baseline}. While average performance is similar across tasks, there are stark differences in DEI metrics. The diversity metric helps discern whether the change in performance is more skewed towards languages with a relatively high or low speaker population. For example, for POS, $\mathrm{MuRIL_{base}}$ and $\mathrm{XLMR_{large}}$ have a 4.2\% difference in performance but a 5.9\% difference in $\mathrm{M_{\tau = 1}}$. This indicates that the difference is more pronounced for languages with large speaker populations. Similarly, for NLI, the difference in performance and $\mathrm{M_{\tau = 1}}$ is 4.6\% and 1.1\% respectively, which also highlights a lack of test data to quantify larger differences in diversity. 
With regards to equity, we observe that even though major differences exist compared to average performance, the Gini coefficient remains relatively unchanged, indicating that while overall performance has increased, the disparity in performance amongst languages has not yet been addressed by any model. Regarding efficiency or inclusion, while $\mathrm{MuRIL_{large}}$ beats $\mathrm{MuRIL_{base}}$ in performance, $\mathrm{MuRIL_{base}}$ is more efficient to use, across all tasks. \\

\noindent \textbf{Where are we today w.r.t DEI of NLP technology?} We report results of best-performing models (fine-tuned on EN and HI) in \reftbl{table:zero_shot_cut} (detailed results with $\mathrm{XLMR_{base}}$ and IndicBERT in \reftbl{table:zero_shot}). Overall, the diversity metric is highest for $\mathrm{MuRIL_{large}}$, when fine-tuned on HI. We also observe that the diversity metric increases with region-specific choices, both in pre-training and fine-tuning. The Gini coefficient remains relatively high at around 0.76 even for the best models, which highlights the disparity in performance even among languages within a single region.\footnote{For comparison, for OECD countries from 2008--2009, the Gini coefficient on income for the entire population ranged between 0.34 and 0.53 while the Gini coefficient for the entire world has been estimated to be between 0.61 and 0.68 \cite{hillebrand2009poverty,klugman2010real}.} With regards to efficiency, averaging across languages and tasks, $\mathrm{MuRIL_{base}}$ performs best. \\
\vspace{-2mm}

\noindent \textbf{What is the way forward?} Overall, the absolute values of the global metric and the Gini coefficient indicate that there lies great potential in both increasing the utility of our models and making them more equitable. Since model performances partially reflect the amount of raw data used in pre-training \cite{lauscher2020zero}, creating equitable unlabeled data resources would alleviate these issues. However, this is an ambitious undertaking that is extremely resource intensive and can certainly not be achieved for 6500 languages in the near future. We thus investigate how limited amounts of data can be used to maximally improve utility and equity during fine-tuning.


\begin{table}[t]
\scalebox{0.57}{
\begin{tabular}{c c c c c c c}
\toprule
\multirow{2}{*}{Metric} & \multirow{2}{*}{Budget} & \multirow{2}{*}{Model} & \multicolumn{4}{c}{Fine-tuning Strategy} \\
& & & English & Hindi & Egalitarian & Greedy \\ \midrule
\multirow{6}{*}{$\mathrm{M_{\tau = 1} \boldsymbol{\uparrow}}$} & \multirow{2}{*}{1,000} & $\mathrm{XLM}$-$\mathrm{R_{large}}$ & 54.0 & 66.2 & 65.4 & 65.3 \\
& & $\mathrm{MuRIL_{large}}$ & 60.4 & 71.3 & \textbf{74.1} & 73.6 \\
& \multirow{2}{*}{5,000} & $\mathrm{XLM}$-$\mathrm{R_{large}}$ & 59.4 & 74.4 & 75.4 & 75.7 \\
& & $\mathrm{MuRIL_{large}}$ & 65.4 & 74.8 & 78.2 & \textbf{78.3} \\
& \multirow{2}{*}{10,000} & $\mathrm{XLM}$-$\mathrm{R_{large}}$ & 59.0 & - & 77.6 & 77.6 \\
& & $\mathrm{MuRIL_{large}}$ & 70.5 & - & 79.6 & \textbf{79.9} \\
\midrule
\multirow{6}{*}{Gini Coeff. $\mathrm{\boldsymbol{\downarrow}}$} & \multirow{2}{*}{1,000} & $\mathrm{XLM}$-$\mathrm{R_{large}}$ & 0.6 & 0.6 & 0.59 & 0.59 \\
& & $\mathrm{MuRIL_{large}}$ & 0.6 & 0.6 & \textbf{0.58} & \textbf{0.58} \\
& \multirow{2}{*}{5,000} & $\mathrm{XLM}$-$\mathrm{R_{large}}$ & 0.6 & 0.59 & 0.59 & 0.59 \\
& & $\mathrm{MuRIL_{large}}$ & 0.59 & 0.59 & \textbf{0.58} & \textbf{0.58} \\
& \multirow{2}{*}{10,000} & $\mathrm{XLM}$-$\mathrm{R_{large}}$ & 0.61 & - & 0.59 & 0.59 \\
& & $\mathrm{MuRIL_{large}}$ & 0.59 & - & \textbf{0.58} & \textbf{0.58} \\
\bottomrule
\end{tabular}}
\caption{\emph{Performance on NER under different annotation budgets}. We observe that the greedy approach (\textsection \ref{sec:greedy}) performs best across all metrics. Note that the HI train set has 5,000 examples only. Details in \textsection \ref{few_shot}.}
\label{table:different_budgets}
\end{table}

\subsection{Few-shot results}
\label{few_shot}

\paragraph{Problem Formulation} For few-shot fine-tuning, we focus on NER where sufficient labeled training data for seven IN languages is available. We employ the source languages $\mathrm{S=\{bn, en, hi, ml, mr, ta, ur\}}$ and seek to optimize metrics on the target languages $\mathrm{T=\{bn, en, gu, hi, ml, mr, pa, ta, te, ur\}}$. In each setting, we have a limited annotation budget, which we can divide among the source languages. We compare against several competitive baselines: \textbf{i)} using only examples from EN or HI respectively; \textbf{ii)} distributing the annotation budget in an egalitarian (uniform) way across all source languages \cite{debnath2021towards}; and \textbf{iii)} our novel greedy approach proposed in \textsection \ref{sec:greedy}. For the greedy approach, we illustrate the best-fit curves for each $\mathrm{(s, t)}$ pair in \refapp{sec:budget_alloc} (\reftbl{table:power_law}). 

\vspace{-1mm}
\paragraph{Results} We show the results under various annotation budgets in \reftbl{table:different_budgets}. Overall, we find that our method yields a higher global metric under most budgets (5 of 6 cases) and also yields a lower Gini coefficient under \emph{all} budget schemes. The optimal allocations for each budget are shown in \reftbl{table:optimal_alloc}. As we can see, the greedy algorithm converges to a solution that is close to uniform. This provides further evidence for the benefits of an egalitarian distribution of annotation budget in order to maximize performance across all languages as the expected marginal gain for languages that have been under-represented during training will be highest. Both the egalitarian and greedy approaches significantly outperform fine-tuning on EN or HI. For instance, our greedy approach outperforms fine-tuning on 10,000 EN examples by 1--3\% with a budget of only 1,000 examples.

\section{Discussion}
\paragraph{Building evaluation datasets} Having uncovered the linguistic inequity and exclusivity of current NLP technologies, we seek to identify practical measures we can take in order to mitigate these biases. As a first step, it is paramount to build representative evaluation sets for all languages as they are required to accurately measure diversity and equity. Out of the 23 languages in our case study, most do not have evaluation data across tasks despite holding official recognition and being spoken by 97\% of the population. In light of the benefits of an egalitarian data distribution during few-shot learning, we also recommend the collection of small amounts of data across many languages for training, in order to maximize marginal gain. These datasets should be collected at the grass-roots level, involving the community they need to serve to capture culturally relevant phenomenon. A prime example of this is the Masakhane organisation\footnote{\href{https://www.masakhane.io/}{https://www.masakhane.io/}} steering efforts towards data collection in African languages, involving the local community. Incentivizing rural, low-income workers to provide for such data also serves as a viable source of supplementary income, and does not degrade dataset quality \cite{abraham2020crowdsourcing}.
\vspace{-1mm}
\paragraph{Trading off multilinguality and regionality} From a modeling perspective, multilingual pre-trained models have been instrumental to NLP systems supporting an unprecedented number of languages, because of their zero-shot transfer capabilities. However, while these are a big step towards linguistic inclusion, they are subject to limitations such as highly skewed pre-training distributions and limited transfer to under-represented languages \cite{Hu2020xtreme, lauscher2020zero}, a bias towards the source language, and sub-optimal tokenization \cite{Wang2021}.
A way to combat these issues is to make region-specific choices, both in pre-training and fine-tuning, as observed in \textsection \ref{sec:zero_shot}. Localizing the problem also enables one to incorporate linguistic expertise \cite{nzeyimana-niyongabo-rubungo-2022-kinyabert} and provide support for culturally relevant phenomena like transliteration or code-mixing. Despite this, we must be wary of excessive fragmentation in pre-training as it leads to higher maintenance costs and there is a possibility that these benefits will be overcome with advances in compute and model capacity in the near future. Optimal fine-tuning however, is promising, as evidenced in \textsection \ref{few_shot} where we observe significant gains in moving away from the zero-shot paradigm. 

\section{Conclusion}

We have proposed a framework for the evaluation of NLP technology based on diversity, equity, and inclusion and proposed the Gini coefficient to quantify equity. We have assessed to what extent several modeling and data choices affect the value NLP technology confers to speakers of Indian languages.
We have also proposed an algorithmic method for resource allocation for task-specific fine-tuning, which outperforms a purely egalitarian distribution of data labeling. Finally, we highlight the importance of building representative evaluation sets from the grass-roots level to enable tracking progress, and discuss how even with the best modeling strategies, we have a long road ahead in building inclusive, equitable systems. While region-specific choices help to a certain extent, building a single global multilingual model without compromising on the three metrics is something we should move towards in the future. We sincerely hope our evaluation paradigm aids in tracking the community's progress in building linguistically diverse technologies.

\section*{Limitations}
We do not consider the inequalities that may exist within subgroups in a language given the lack of fine-grained evaluation data. In multilingual countries like India, each language is composed of several dialects (Hindi alone is composed of 58 dialects \cite{india2011census}). As disparities exist along multiple axes such as caste, gender, religion and so on \cite{sambasivan2021re}, it is imperative to go beyond the language level. We only consider pre-trained language models for our experiments given their massive language coverage and zero-shot transfer capabilities. There have been efforts to build language-specific, task-specific models which we do not include in our study. Our greedy data allocation method is a strong baseline that outperforms standard approaches such as selecting a single source language or uniform selection. It can be improved by incorporating notions of language similarity, which requires more complex methods \cite{Song2019}.

\section*{Acknowledgements}

We would like to thank the reviewers for their insightful feedback. We would also like to thank Melvin Johnson and Slav Petrov for helpful feedback on a draft of this post.


\bibliography{emnlp2022}
\bibliographystyle{acl_natbib}

\appendix


\section{Appendix}
\label{sec:appendix}
\subsection{Efficiency}
\label{appendix:efficiency}
We report the throughput and memory for each model and task in \reftbl{table:efficiency_metrics}. For NLI, POS and NER, the maximum sequence length is 128 and for QA it's 384.

As detailed in \refsec{sec:efficiency}, we need to calculate the $\mathrm{AMRS}$ for each metric $\mathrm{M}$ (throughput and memory saved), to calculate the efficiency score. As described in \cite{ma2021dynaboard}, each model has properties (or \emph{goods}) that inform its utility. Here, these goods are throughput, memory saved, and performance. A model is a point in this space of goods and an indifference curve is a set of points that provide the same utility (for different values of these properties). These curves are monotonically negatively sloped, i.e., for a model with higher accuracy to be on the same curve as one with a lower accuracy, it will have to use up more memory or have lower throughput. For a given indifference curve, the rate at which this trade-off is made, is called the marginal rate of substitution ($\mathrm{MRS}$). 

To calculate $\mathrm{MRS}$, and consequently $\mathrm{AMRS}$, \cite{ma2021dynaboard} make two key assumptions: \textbf{i)} All models lie on the same indifference curve; \textbf{ii)} if $\mathrm{M(x_i) > M(x_{i+1})}$ and $\mathrm{perf(x_i) > perf(x_{i+1})}$, then there exists a model $\mathrm{\left< perf(x_{i+1}), M(x_i) + (M(x_i) - M(x_{i+1}))\right>}$ on the same indifference curve as $\mathrm{x_i}$. For our case study, we believe that assuming regional and global models to lie on the same indifference curve would be inaccurate, since models with the same capacity (size and architecture) have been trained on a different set of languages. In the case of \cite{ma2021dynaboard}, they only consider models pre-trained on English. Here, we assume that regional models (trained on 15-17 languages) would be strictly better on all dimensions and hence lie on a different indifference curve as compared to global models (trained on 100+ languages). Hence, we assume IndicBERT, $\mathrm{MuRIL_{base}}$ and $\mathrm{MuRIL_{large}}$ to lie on one indifference curve and $\mathrm{XLMR_{base}}$ and $\mathrm{XLMR_{large}}$ to lie on another. The second assumption holds in our case as well.

For a model $\mathrm{x_i}$, $\mathrm{Efficiency(x_i)}$, $\mathrm{MRS}$, and $\mathrm{AMRS}$ are given by : 
\begin{align*}
& \mathrm {Efficiency(x_i) = \sum_{M} w_{M} * \frac{M(x_i)}{AMRS(M, perf)}} \\
& \mathrm {AMRS(M, perf) = \overline{MRS}} \\
& \mathrm {MRS = \Bigg\{ \Bigg| \frac{M(x_i) - M(x_{i+1})}{perf(x_i) - perf(x_{i+1})}\Bigg| 1 \leq i < n \Bigg\}}
\end{align*}


\begin{table}[ht]
\scalebox{0.75}{
\begin{tabular}{c | c | c | c | c | c }
\hline Model & Metric & NER & PoS & NLI & QA \\
\hline \multirow{3}{*}{IndicBERT} & Memory Saved & \multicolumn{4}{c}{15.9GB} \\
\cdashline{2-6} & Throughput & 22.6 & 20.2 & 22.9 & 10.5 \\
\cdashline{2-6} & Perf (EN) & 41.3 & 71.6 & 69.7 & 52.4 \\
\hline \multirow{3}{*}{$\mathrm{XLM}$-$\mathrm{R_{base}}$} & Memory Saved & \multicolumn{4}{c}{15GB} \\
\cdashline{2-6} & Throughput & 24.4 & 23.2 & 26.4 & 14.9 \\
\cdashline{2-6} & Perf (EN) & 61.7 & 82.2 & 77.1 & 72.1 \\
\hline \multirow{3}{*}{$\mathrm{MuRIL_{base}}$} & Memory Saved & \multicolumn{4}{c}{15.1GB} \\
\cdashline{2-6} & Throughput & 23.8 & 23.1 & 26.2 & 15.7 \\
\cdashline{2-6} & Perf (EN) & 74.9 & 80.3 & 78.9 & 77.5 \\
\hline \multirow{3}{*}{$\mathrm{XLM}$-$\mathrm{R_{large}}$} & Memory Saved & \multicolumn{4}{c}{13.9GB} \\
\cdashline{2-6} & Throughput & 9.4 & 10.0 & 10.4 & 4.1 \\
\cdashline{2-6} & Perf (EN) & 64.6 & 83.7 & 81.8 & 81.4 \\
\hline \multirow{3}{*}{$\mathrm{MuRIL_{large}}$} & Memory Saved & \multicolumn{4}{c}{14.1GB} \\
\cdashline{2-6} & Throughput & 9.8 & 9.9 & 10.5 & 4.2 \\
\cdashline{2-6} & Perf (EN) & 71.8 & 83.4 & 82.8 & 83.0 \\
\hline \multirow{2}{*}{AMRS (Regional)} & Throughput & 1.7 & 3.4 & 2.2 & 1.1 \\
\cdashline{2-6} & Memory & 0.1 & 0.3 & 0.2 & 0.1 \\
\hline \multirow{2}{*}{AMRS (Global)} & Throughput & 3.7 & 7.1 & 3.3 & 1.2 \\
\cdashline{2-6} & Memory & 0.3 & 0.5 & 0.2 & 0.1 \\
\hline
\end{tabular}}
\caption{\label{table:efficiency_metrics} The throughput is given by the number of instances processed per second by the fine-tuned models on CPU.}
\end{table}

\subsection{Pre-training Languages}
\label{app:pretraining}
In \refsec{sec:setup}, we choose IndicBERT, MuRIL and XLM-R as pre-trained multilingual models to base our analysis upon. IndicBERT is trained on 11 IN languages that include Assamese (as), Bengali (bn), Gujarati (gu), Hindi (hi), Kannada (kn), Malayalam (ml), Marathi (mr), Oriya (or), Punjabi (pa), Tamil (ta), Telugu (te). XLM-R includes 15 IN languages in training with the addition of Nepali (ne), Sanskrit (sa), Sindhi (sd) and Urdu (ur) over IndicBERT and MuRIL is trained on 16 IN languages, with the addition of Kashmiri (ks) over XLM-R.

\subsection{Gini Coefficient}
\label{app:gini}

The Gini coefficient is mathematically computed based on the Lorenz curve, which plots the relation between population size and the cumulative income earned by that population as shown in \reffig{fig:gini}. To plot the Lorenz curve, individuals are sorted in increasing order of income ($\mathrm{x}$-axis) and their cumulative wealth is plotted on the $\mathrm{y}$-axis. In essence, a point ($\mathrm{x}$, $\mathrm{y}$) indicates that the bottom $\mathrm{x}$\% of the population holds $\mathrm{y}$ amount of wealth. The line at 45 degrees represents perfect equality of incomes. The Gini coefficient $\mathrm{G}$ is then calculated as the ratio of the area that lies between the line of equality and the Lorenz curve ($\mathrm{A}$ in \reffig{fig:gini}), over the total area under the line of equality ($\mathrm{A + B}$ in \reffig{fig:gini}). If
$\mathrm{G = 0}$,  every person in the population receives an equal percentage of income and if $\mathrm{G = 1}$, a single person receives 100\% of the income. Since the axes scale from 0 to 1, $\mathrm{A + B = 0.5}$. In essence, if the Lorenz curve is represented by the function $\mathrm{Y = L(X)}$ then $\mathrm{G}$ can be given as:

\vspace{-5mm}
{\small
\begin{align*}
\mathrm {G = \frac{A}{A + B} = 2A = 1-2B = 1-2 \int_{0}^{1} L(X)dX}
\end{align*}
}%

\noindent For a population with values $\mathrm{y_i}$, $\mathrm{i = 1\ ...\ n}$, that are indexed in non-decreasing order ($\mathrm{y_i \leq y_{i+1}}$):
\begin{align*}
\mathrm {G = \frac{1}{n}\left(n+1 - 2\frac{\sum_{i=1}^{n} (n+1-i)y_i}{\sum_{i=1}^{n}y_i}\right)}
\end{align*}

For comparison, for OECD countries from 2008--2009, the Gini coefficient on income for the entire population ranged between 0.34 and 0.53. The Gini coefficient on income for the entire world has been estimated to be between 0.61 and 0.68 \cite{hillebrand2009poverty,klugman2010real}. In our experiments on Indian languages, state-of-the-art models achieve an average Gini coefficient of 0.77, which highlights the disparity in performance even among languages within a single region.

As mentioned in \refsec{sec:zero_shot}, calculating the Gini coefficient across all 23 languages doesn't reflect the dispersion in performances across languages for which we have test sets. To compare between baselines, we additionally report the Gini coefficient evaluated only across those languages for which we have test sets as shown in \reftbl{table:gini_zero_shot}. We observe that region-specific choices ($\mathrm{MuRIL_{base}}$ fine-tuned on HI) lead to the lowest value, similar to what we observe with the global metric.

\subsection{Fine-tuning Details}
We fine-tune all models using the hyperparameters mentioned in \reftbl{tab:ft_hparams} for each task and model consistently throughout the paper. We make use of the XTREME codebase\footnote{\href{https://github.com/google-research/xtreme}{https://github.com/google-research/xtreme}} to finetune these models using a NVIDIA A100 GPU. We make an exception for IndicBERT when fine-tuning on NER, where we fine-tune for 15 epochs instead of 10, to reach convergence.

\begin{table}[t]
\centering
\scalebox{0.55}{
\begin{tabular}{c | c | c | c | c | c | c | c}
\hline Metric & Train Lang. & Model & NER & PoS & NLI & QA & Average \\
\hline \multirow{10}{*}{Gini Coeff. $\boldsymbol{\downarrow}$} & \multirow{5}{*}{English} & IndicBERT & 0.155 & 0.107 & 0.051 & 0.091 & 0.101 \\
& & $\mathrm{XLM}$-$\mathrm{R_{base}}$ & 0.095 & 0.067 & 0.058 & 0.048 & 0.067 \\
& & $\mathrm{MuRIL_{base}}$ & 0.047 & 0.086 & 0.048 & 0.03 & \textbf{0.052} \\
& & $\mathrm{XLM}$-$\mathrm{R_{large}}$ & 0.084 & 0.06 & 0.049 & 0.026 & 0.055 \\
& & $\mathrm{MuRIL_{large}}$ & 0.051 & 0.086 & 0.051 & 0.027 & 0.057 \\
\cdashline{2-8} & \multirow{5}{*}{Hindi} & IndicBERT & 0.173 & 0.073 & 0.004 & 0.041 & 0.073 \\
& & $\mathrm{XLM}$-$\mathrm{R_{base}}$ & 0.067 & 0.037 & 0.039 & 0.046 & 0.047 \\
& & $\mathrm{MuRIL_{base}}$ & 0.062 & 0.032 & 0.036 & 0.012 & \textbf{0.035} \\
& & $\mathrm{XLM}$-$\mathrm{R_{large}}$ & 0.057 & 0.04 & 0.033 & 0.029 & 0.04 \\
& & $\mathrm{MuRIL_{large}}$ & 0.065 & 0.057 & 0.033 & 0.014 & 0.042 \\
\hline
\end{tabular}}
\caption{\label{table:gini_zero_shot}\emph{Gini Coefficient} for all models calculated only across languages having evaluation sets for each task. }
\end{table}

\begin{table}[t]
\centering
\scalebox{0.75}{
\begin{tabular}{c | c | c | c | c | c }
\hline \multirow{2}{*}{Task} & Batch & Learning & No. of & Warmup & Max. seq. \\ 
& Size & Rate & Epochs & Ratio & Length \\
\hline
NER & 32 & 2e-5 & 10 & 0.1 & 128 \\
POS & 32 & 2e-5 & 10 & 0.1 & 128 \\
NLI & 64 & 2e-5 & 3 & 0.1 & 128 \\
QA & 32 & 3e-5 & 2 & 0.1 & 384 \\
\hline 
\end{tabular}}
\caption{\label{tab:ft_hparams} Hyperparameter details for each fine-tuning task}
\end{table}

\subsection{Budget Allocation}
\label{sec:budget_alloc}
\begin{figure*}[t]
\includegraphics[scale=0.2]{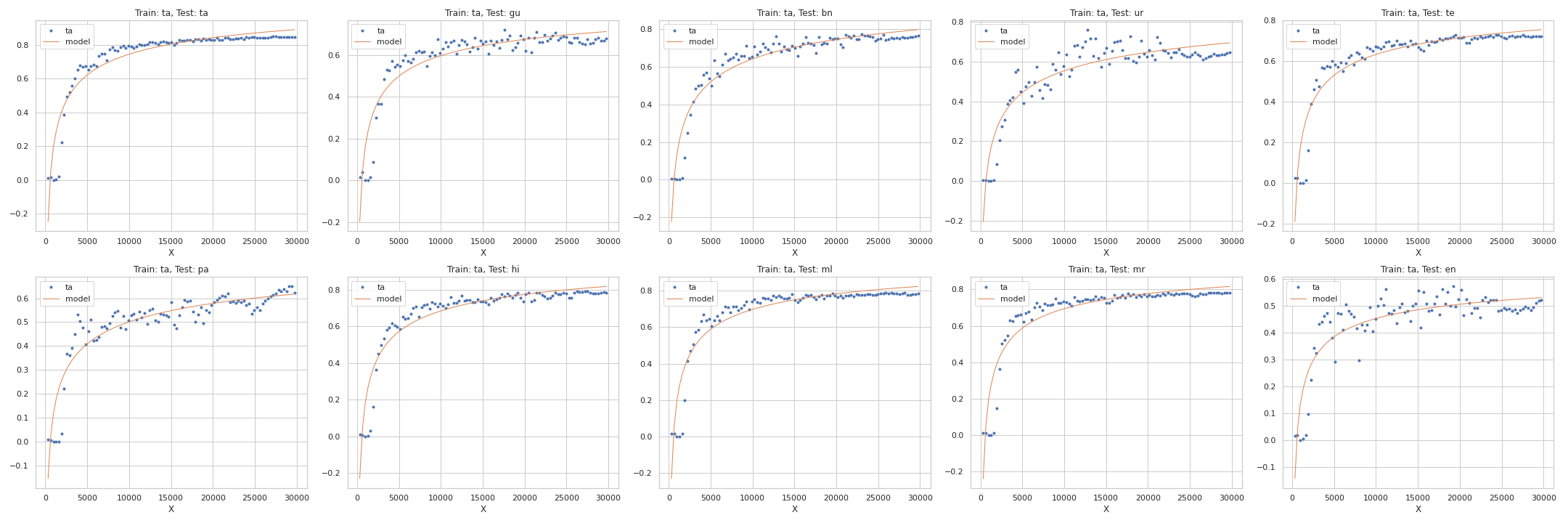} 
\caption{\emph{Best-fit curves for XLM-R} when fine-tuned on Tamil for each of the target languages.}
\label{fig:best_fit_xlmr}
\end{figure*}

\begin{figure*}[t]
\includegraphics[scale=0.2]{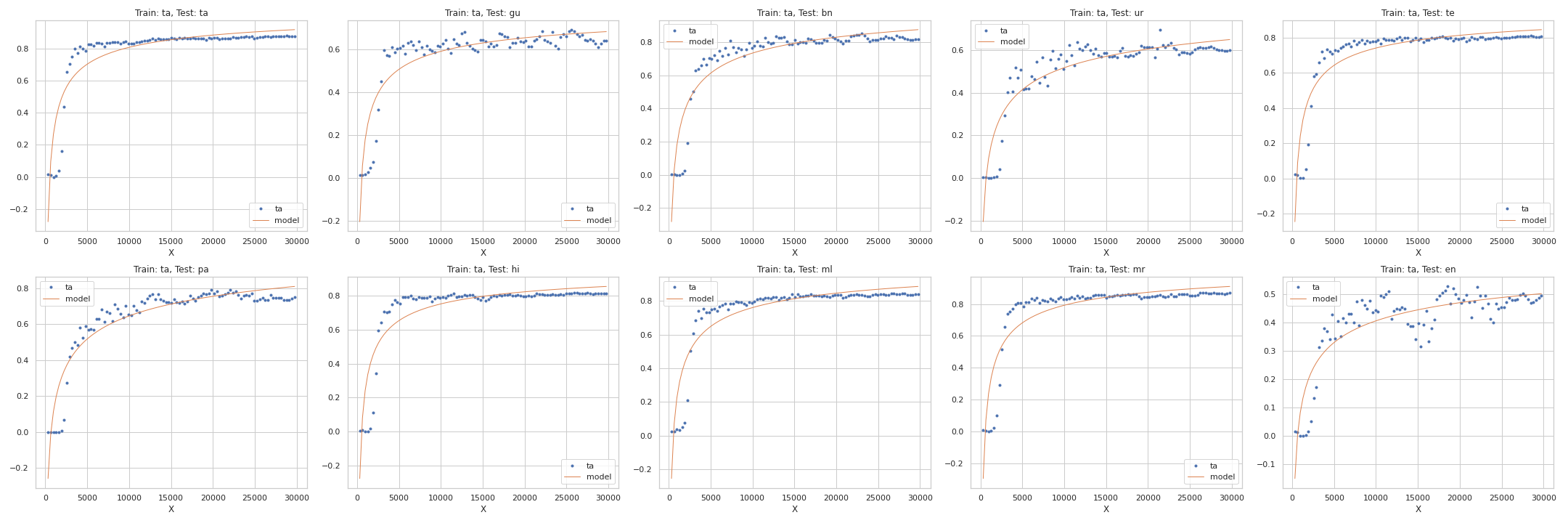} 
\caption{\emph{Best-fit curves for MuRIL} when fine-tuned on Tamil for each of the target languages.}
\label{fig:best_fit_muril}
\end{figure*}

In \refsec{sec:greedy}, we describe an empirical budget allocation scheme for fine-tuning of pre-trained models that can jointly optimize on our proposed metrics. We follow a greedy approach to solve this problem, as shown in \reftbl{algorithm_1}. In this paper, we solve this for one task, namely NER, but the methodology proposed is generally extensible to any task and combination of languages since it is purely empirical. We select seven source languages for which we have enough training data and fine-tune $\mathrm{MuRIL_{large}}$ and $\mathrm{XLM}$-$\mathrm{R_{large}}$ for each of these source languages independently, for two epochs. During fine-tuning, we evaluate on each of our target languages after every 10 steps of training. Given our batch-size is 32, we gather data-points at a step size of 320 training instances. Consequently, say we have 5000 training instances for a source language, we gather approximately 30 sample points for that source language and any target language. Using these, we plot best-fit curves for $\mathrm{\forall (s, t)}$ pairs using the \emph{scipy.optimize.curve\_fit} package. Given a function, $\mathrm{f(x)}$, \emph{curve\_fit} uses non-linear least squares to fit $\mathrm{f(x)}$ to the observed data-points. We define $\mathrm{f(x)_{s, t} = a_{s, t} + b_{s, t} * x^{-c_{s, t}}}$, because the relation between model performance and training data follows a power-law distribution \cite{rosenfeld2019constructive}. The best-fit curves for each source and target pair are shown in \reftbl{table:power_law}. The visualizations of the best-fit curves for a sample training language (Tamil) are shown in Figures \ref{fig:best_fit_xlmr}, \ref{fig:best_fit_muril}. Having determined constant values \{$\mathrm{a_{s, t}, b_{s, t},  c_{s, t}}$\} $\mathrm{\forall (s, t)}$ independently, we proceed with finding the optimal allocation using the algorithm described in \reftbl{algorithm_1}. We solve this for three different budgets, i.e., 1,000; 5,000 and 10,000 and the optimal allocations for each budget are shown in \reftbl{table:optimal_alloc}. 

\begin{table*}[t]
\centering
\scalebox{0.7}{
\begin{tabular}{p{0.02\linewidth} p{0.98\linewidth}}
\hline \multicolumn{2}{l}{\textbf{Greedy Algorithm}} \\
\hline 1: & \textbf{Input}: Fine-tuning labeled data $\mathrm{\forall s\ \epsilon\ S}$. A fixed budget of labeled data instances $\mathrm{X}$ \\
2: & \textbf{Initialize}: Set the total number of allocated instances to zero, i.e., $\mathrm{allocated = 0}$, the number of allocated samples for each source language to zero, i.e. $\mathrm{samples[s]=0 \forall s\ \epsilon\ S}$, the current global metric for each source language to -inf, i.e. $\mathrm{current\_gm[s]=-inf \forall s\ \epsilon\ S}$ and the current gini coefficient for each source language to 1, i.e. $\mathrm{current\_gini[s]=1 \forall s\ \epsilon\ S}$ \\
3: & \textbf{while} $\mathrm{allocated < X}$ \textbf{do} \\
4: & \hspace{3mm} $\mathrm{highest\_marginal\_gain = 0}$ \\
5: & \hspace{3mm} \textbf{for} $\mathrm{s}$ in $\mathrm{S}$ \textbf{do} \\
6: & \hspace{6mm} $\mathrm{gm_{s} = \sum_{t \epsilon T}\ d_t^{(\tau)}* (a_{s,t} + b_{s,t} * (samples[s] + 1)^{-c_{s,t}})}$ \\
7: & \hspace{6mm} $\mathrm{gini_{s} = F[abs(performance_{s, t}(samples[s] + 1)) \forall t\ \epsilon\ T ]}$ \\
8: & \hspace{6mm} $\mathrm{\Delta gm_{s} = gm_{s} - current\_gm[s]}$ \\
9: & \hspace{6mm} $\mathrm{\Delta gini_{s} = current\_gini[s] - gini_{s}}$ \\
10: & \hspace{6mm} $\mathrm{marginal\_gain = \alpha * \Delta gm_{s} + \beta * \Delta gini_{s}}$  \\
9: & \hspace{6mm} \textbf{if} $\mathrm{marginal\_gain > highest\_marginal\_gain}$ \textbf{do} \\
10: & \hspace{9mm} $\mathrm{highest\_marginal\_gain = marginal\_gain}$ \\
11: & \hspace{9mm} $\mathrm{best\_language = s}$ \\
12: & \hspace{9mm} $\mathrm{best\_gm = gm_{s}}$ \\
13: & \hspace{9mm} $\mathrm{best\_gini = gini_{s}}$ \\
14: & \hspace{6mm} \textbf{end if} \\
15: & \hspace{3mm} \textbf{end for} \\
16: & \hspace{3mm} $\mathrm{samples[s] = samples[s] + 1}$ \\
17: & \hspace{3mm} $\mathrm{allocated = allocated + 1}$ \\
18: & \hspace{3mm} $\mathrm{current\_gm[best\_language] = best\_gm}$ \\
19: & \hspace{3mm} $\mathrm{current\_gini[best\_language] = best\_gini}$ \\
20: &\textbf{end while} \\
\hline
\end{tabular}}
\caption{\label{algorithm_1}A greedy approach to solve the constrained optimization for the budget allocation problem as described in \refapp{sec:budget_alloc}.}
\end{table*}

\begin{table}[t]
\scalebox{0.6}{
\begin{tabular}{c | c | c | c | c | c }
\hline \multirow{2}{*}{Test} & \multirow{2}{*}{Train} & \multicolumn{2}{c|}{MuRIL} & \multicolumn{2}{c}{XLM-R} \\
\cline{3-6} & & Edge Weight & R-squared & Edge Weight & R-squared \\
\hline \multirow{7}{*}{bn} & bn & $\mathrm{1.2-29.0*x^{-0.5}}$ & 0.88 & $\mathrm{1.3-11.5*x^{-0.4}}$ & 0.93 \\ 
& en & $\mathrm{1.2-11.4*x^{-0.4}}$ & 0.78 & $\mathrm{1.1-8.1*x^{-0.3}}$ & 0.89 \\ 
& hi & $\mathrm{1.4-9.4*x^{-0.3}}$ & 0.85 & $\mathrm{1.1-8.1*x^{-0.3}}$ & 0.92 \\
& ml & $\mathrm{1.2-10.7*x^{-0.3}}$ & 0.86 & $\mathrm{2.3-4.8*x^{-0.1}}$ & 0.92 \\
& mr & $\mathrm{1.9-6.5*x^{-0.2}}$ & 0.88 & $\mathrm{1.9-4.6*x^{-0.1}}$ & 0.93 \\
& ta & $\mathrm{1.2-10.5*x^{-0.3}}$ & 0.83 & $\mathrm{1.3-6.1*x^{-0.2}}$ & 0.90 \\
& ur & $\mathrm{1.0-13.5*x^{-0.4}}$ & 0.88 & $\mathrm{1.0-6.5*x^{-0.3}}$ & 0.91 \\
\hline \multirow{7}{*}{en} & bn & $\mathrm{0.9-4.4*x^{-0.3}}$ & 0.86 & $\mathrm{1.0-5.2*x^{-0.3}}$ & 0.90 \\ 
& en & $\mathrm{1.1-16.4*x^{-0.4}}$ & 0.82 & $\mathrm{1.1-14.6*x^{-0.4}}$ & 0.85 \\ 
& hi & $\mathrm{1.0-5.6*x^{-0.3}}$ & 0.88 & $\mathrm{1.0-7.6*x^{-0.3}}$ & 0.90 \\
& ml & $\mathrm{1.9-3.5*x^{-0.1}}$ & 0.88 & $\mathrm{1.0-6.1*x^{-0.3}}$ & 0.86 \\
& mr & $\mathrm{1.2-3.2*x^{-0.2}}$ & 0.84 & $\mathrm{1.2-4.8*x^{-0.2}}$ & 0.91 \\
& ta & $\mathrm{0.8-4.2*x^{-0.3}}$ & 0.76 & $\mathrm{0.7-6.9*x^{-0.4}}$ & 0.76 \\
& ur & - & 0.88 & $\mathrm{1.0-3.9*x^{-0.2}}$ & 0.90 \\
\hline \multirow{7}{*}{gu} & bn & $\mathrm{2.6-4.3*x^{-0.1}}$ & 0.93 & $\mathrm{1.0-4.8*x^{-0.3}}$ & 0.88 \\ 
& en & $\mathrm{0.9-5.5*x^{-0.3}}$ & 0.80 & $\mathrm{0.7-11.3*x^{-0.5}}$ & 0.78 \\ 
& hi & $\mathrm{1.2-5.4*x^{-0.2}}$ & 0.87 & $\mathrm{0.7-13.3*x^{-0.5}}$ & 0.86 \\
& ml & $\mathrm{1.2-7.8*x^{-0.3}}$ & 0.85 & $\mathrm{1.6-4.3*x^{-0.2}}$ & 0.90 \\
& mr & $\mathrm{1.1-6.4*x^{-0.3}}$ & 0.87 & $\mathrm{1.1-6.0*x^{-0.3}}$ & 0.85 \\
& ta & $\mathrm{0.8-11.3*x^{-0.4}}$ & 0.78 & $\mathrm{1.0-7.6*x^{-0.3}}$ & 0.84 \\
& ur & $\mathrm{1.4-3.4*x^{-0.1}}$ & 0.91 & $\mathrm{1.6-3.2*x^{-0.1}}$ & 0.89 \\
\hline \multirow{7}{*}{hi} & bn & $\mathrm{0.9-17.2*x^{-0.5}}$ & 0.88 & $\mathrm{1.2-4.8*x^{-0.2}}$ & 0.94 \\ 
& en & $\mathrm{1.0-11.7*x^{-0.4}}$ & 0.83 & $\mathrm{0.9-8.6*x^{-0.4}}$ & 0.88 \\ 
& hi & $\mathrm{1.1-23.9*x^{-0.5}}$ & 0.90 & $\mathrm{1.3-9.5*x^{-0.3}}$ & 0.92 \\
& ml & $\mathrm{1.1-12.4*x^{-0.4}}$ & 0.85 & $\mathrm{1.4-5.7*x^{-0.2}}$ & 0.90 \\
& mr & $\mathrm{1.1-17.6*x^{-0.5}}$ & 0.85 & $\mathrm{2.0-5.3*x^{-0.2}}$ & 0.93 \\
& ta & $\mathrm{1.0-19.1*x^{-0.5}}$ & 0.78 & $\mathrm{1.1-8.7*x^{-0.3}}$ & 0.88 \\
& ur & $\mathrm{1.0-8.0*x^{-0.3}}$ & 0.92 & $\mathrm{1.2-4.6*x^{-0.2}}$ & 0.94 \\
\hline \multirow{7}{*}{ml} & bn & $\mathrm{1.1-5.9*x^{-0.3}}$ & 0.88 & $\mathrm{1.3-4.3*x^{-0.2}}$ & 0.92 \\ 
& en & $\mathrm{1.2-5.0*x^{-0.2}}$ & 0.85 & $\mathrm{0.8-7.6*x^{-0.3}}$ & 0.85 \\ 
& hi & $\mathrm{2.1-5.0*x^{-0.1}}$ & 0.86 & $\mathrm{1.0-10.8*x^{-0.4}}$ & 0.90 \\
& ml & $\mathrm{1.1-21.4*x^{-0.5}}$ & 0.83 & $\mathrm{1.3-7.8*x^{-0.3}}$ & 0.90 \\
& mr & $\mathrm{1.5-7.3*x^{-0.3}}$ & 0.86 & $\mathrm{1.4-6.4*x^{-0.3}}$ & 0.91 \\
& ta & $\mathrm{1.1-12.8*x^{-0.4}}$ & 0.81 & $\mathrm{1.1-10.0*x^{-0.4}}$ & 0.86 \\
& ur & $\mathrm{1.1-5.1*x^{-0.2}}$ & 0.89 & $\mathrm{1.1-4.7*x^{-0.2}}$ & 0.89 \\
\hline \multirow{7}{*}{mr} & bn & $\mathrm{1.0-9.3*x^{-0.4}}$ & 0.88 & $\mathrm{1.1-5.1*x^{-0.2}}$ & 0.89 \\ 
& en & $\mathrm{0.9-8.8*x^{-0.3}}$ & 0.81 & $\mathrm{0.9-7.6*x^{-0.3}}$ & 0.87 \\ 
& hi & $\mathrm{1.3-10.3*x^{-0.3}}$ & 0.86 & $\mathrm{1.1-9.6*x^{-0.4}}$ & 0.91 \\
& ml & $\mathrm{1.1-15.2*x^{-0.4}}$ & 0.83 & $\mathrm{1.3-6.4*x^{-0.3}}$ & 0.90 \\
& mr & $\mathrm{1.2-21.9*x^{-0.5}}$ & 0.86 & $\mathrm{1.6-7.6*x^{-0.3}}$ & 0.92 \\
& ta & $\mathrm{1.1-17.3*x^{-0.4}}$ & 0.79 & $\mathrm{1.1-10.7*x^{-0.4}}$ & 0.85 \\
& ur & $\mathrm{1.2-5.2*x^{-0.2}}$ & 0.92 & $\mathrm{1.3-4.2*x^{-0.2}}$ & 0.91 \\
\hline \multirow{7}{*}{pa} & bn & $\mathrm{1.0-6.4*x^{-0.3}}$ & 0.86 & $\mathrm{0.9-4.2*x^{-0.3}}$ & 0.82 \\ 
& en & $\mathrm{1.2-4.0*x^{-0.2}}$ & 0.84 & $\mathrm{1.1-2.9*x^{-0.2}}$ & 0.85 \\ 
& hi & $\mathrm{1.9-5.9*x^{-0.2}}$ & 0.84 & $\mathrm{1.8-4.0*x^{-0.1}}$ & 0.93 \\
& ml & $\mathrm{1.0-9.7*x^{-0.4}}$ & 0.83 & $\mathrm{1.2-3.6*x^{-0.2}}$ & 0.87 \\
& mr & $\mathrm{2.7-5.3*x^{-0.1}}$ & 0.88 & $\mathrm{1.3-3.9*x^{-0.2}}$ & 0.84 \\
& ta & $\mathrm{1.4-6.3*x^{-0.2}}$ & 0.86 & $\mathrm{1.0-4.7*x^{-0.2}}$ & 0.84 \\
& ur & $\mathrm{1.2-4.5*x^{-0.2}}$ & 0.92 & $\mathrm{0.8-4.2*x^{-0.3}}$ & 0.87 \\
\hline \multirow{7}{*}{ta} & bn & $\mathrm{1.1-7.2*x^{-0.3}}$ & 0.89 & $\mathrm{1.0-4.6*x^{-0.2}}$ & 0.93 \\ 
& en & $\mathrm{1.0-7.9*x^{-0.3}}$ & 0.83 & $\mathrm{0.8-6.7*x^{-0.3}}$ & 0.86 \\ 
& hi & $\mathrm{1.4-6.7*x^{-0.3}}$ & 0.90 & $\mathrm{1.2-5.9*x^{-0.3}}$ & 0.92 \\
& ml & $\mathrm{1.0-14.1*x^{-0.4}}$ & 0.83 & $\mathrm{1.3-4.7*x^{-0.2}}$ & 0.92 \\
& mr & $\mathrm{1.3-9.7*x^{-0.3}}$ & 0.86 & $\mathrm{2.7-5.0*x^{-0.1}}$ & 0.94 \\
& ta & $\mathrm{1.1-19.7*x^{-0.5}}$ & 0.79 & $\mathrm{1.2-9.4*x^{-0.3}}$ & 0.88 \\
& ur & $\mathrm{1.2-5.0*x^{-0.2}}$ & 0.92 & $\mathrm{1.5-3.4*x^{-0.1}}$ & 0.92 \\
\hline \multirow{7}{*}{te} & bn & $\mathrm{1.1-5.4*x^{-0.3}}$ & 0.90 & $\mathrm{0.8-4.7*x^{-0.3}}$ & 0.88 \\ 
& en & $\mathrm{0.8-10.1*x^{-0.4}}$ & 0.79 & $\mathrm{0.7-6.7*x^{-0.4}}$ & 0.83 \\ 
& hi & $\mathrm{1.0-9.7*x^{-0.4}}$ & 0.91 & $\mathrm{0.9-7.3*x^{-0.3}}$ & 0.86 \\
& ml & $\mathrm{1.0-15.3*x^{-0.4}}$ & 0.83 & $\mathrm{1.1-5.6*x^{-0.3}}$ & 0.88 \\
& mr & $\mathrm{1.0-12.5*x^{-0.4}}$ & 0.87 & $\mathrm{1.7-4.5*x^{-0.2}}$ & 0.93 \\
& ta & $\mathrm{1.0-16.5*x^{-0.4}}$ & 0.81 & $\mathrm{1.0-7.7*x^{-0.3}}$ & 0.87 \\
& ur & $\mathrm{1.4-4.2*x^{-0.2}}$ & 0.91 & $\mathrm{1.4-3.1*x^{-0.1}}$ & 0.90 \\
\hline \multirow{7}{*}{ur} & bn & $\mathrm{0.6-11.3*x^{-0.5}}$ & 0.86 & - & 0.76 \\ 
& en & $\mathrm{1.1-5.5*x^{-0.2}}$ & 0.83 & $\mathrm{1.0-5.9*x^{-0.3}}$ & 0.81 \\ 
& hi & $\mathrm{2.6-4.8*x^{-0.1}}$ & 0.85 & - & 0.96 \\
& ml & $\mathrm{1.1-8.2*x^{-0.3}}$ & 0.80 & $\mathrm{2.5-5.0*x^{-0.1}}$ & 0.85 \\
& mr & $\mathrm{4.3-6.1*x^{-0.1}}$ & 0.87 & $\mathrm{5.2-7.0*x^{-0.0}}$ & 0.91 \\
& ta & $\mathrm{1.1-5.0*x^{-0.2}}$ & 0.83 & $\mathrm{1.2-5.2*x^{-0.2}}$ & 0.83 \\
& ur & $\mathrm{1.0-42.2*x^{-0.6}}$ & 0.87 & $\mathrm{1.1-20.9*x^{-0.5}}$ & 0.90 \\
\hline
\end{tabular}}
\caption{\label{table:power_law}\emph{Power-law equations} empirically determined for each source and target pair. Please refer to \refsec{sec:budget_alloc} for more details}
\end{table}

\begin{table}[ht]
\scalebox{0.55}{
\begin{tabular}{c | c | c | c | c | c | c | c | c | c }
\hline Metric & Budget & Model & bn & en & hi & ml & mr & ta & ur \\
\hline \multirow{6}{*}{$\mathrm{GM_{\tau = 0}}$} & \multirow{2}{*}{1,000} & $\mathrm{XLM}$-$\mathrm{R_{large}}$ & 128& 157& 145& 134& 133& 163& 140 \\
& & $\mathrm{MuRIL_{large}}$ &137& 135& 134& 158& 142& 159& 135 \\
\cline{2-10} & \multirow{2}{*}{5,000} & $\mathrm{XLM}$-$\mathrm{R_{large}}$ & 704& 792& 693& 794& 696& 628& 693 \\
& & $\mathrm{MuRIL_{large}}$ & 743 & 644 & 749 & 783 & 745 & 852 & 484 \\
\cline{2-10} & \multirow{2}{*}{10,000} & $\mathrm{XLM}$-$\mathrm{R_{large}}$ & 1322& 1349& 1400& 1481& 1457& 1479& 1512 \\

& & $\mathrm{MuRIL_{large}}$ & 1302 & 1468 & 1379 & 1421 & 1425 & 1448 & 1557 \\
\hline \multirow{6}{*}{$\mathrm{GM_{\tau = 1}}$} & \multirow{2}{*}{1,000} & $\mathrm{XLM}$-$\mathrm{R_{large}}$& 126& 160& 159& 134& 129& 163& 129 \\
& & $\mathrm{MuRIL_{large}}$ &142& 136& 152& 143& 148& 157& 122 \\
\cline{2-10} & \multirow{2}{*}{5,000} & $\mathrm{XLM}$-$\mathrm{R_{large}}$ & 710& 805& 713& 803& 707& 639& 623 \\
& & $\mathrm{MuRIL_{large}}$ &744& 644 & 761 & 772 & 747 & 848 & 484 \\
\cline{2-10} & \multirow{2}{*}{10,000} & $\mathrm{XLM}$-$\mathrm{R_{large}}$ & 1308& 1363& 1456& 1465& 1459& 1471& 1478 \\
& & $\mathrm{MuRIL_{large}}$ & 1308 & 1488 & 1396 & 1406 & 1416 & 1441 & 1545 \\
\hline
\end{tabular}}
\caption{\emph{Optimal allocations under different budgets}. Please refer to \refsec{sec:budget_alloc} for more details}
\label{table:optimal_alloc}
\end{table}

\begin{table}[t]
\centering
\scalebox{0.55}{
\begin{tabular}{c c c c c c c c}
\toprule
Metric & Train Lang. & Model & NER & POS & NLI & QA & Average \\ \midrule
\multirow[b]{5}{*}{$\mathrm{M_{\tau = 0} \boldsymbol{\uparrow}}$} & \multirow{5}{*}{English} & IndicBERT & 16.5 & 16.1 & 6.5 & 5.3 & 11.1 \\
& & $\mathrm{XLM}$-$\mathrm{R_{base}}$ & 27.0 & 21.4 & 10.3 & 13.8 & 18.1 \\
& & $\mathrm{MuRIL_{base}}$ & 33.4 & 20.7 & 10.5 & 14.9 & 19.9 \\
& & $\mathrm{XLM}$-$\mathrm{R_{large}}$ & 28.7 & 21.9 & 11.0 & 15.6 & 19.3 \\
& & $\mathrm{MuRIL_{large}}$ & 31.5 & 21.3 & 11.1 & \textbf{15.9} & 20.0 \\
\cline{2-8} \multirow[t]{5}{*}{(Linguistic)} & \multirow{5}{*}{Hindi} & IndicBERT & 23.7 & 17.6 & 6.6 & 4.8 & 13.2 \\
& & $\mathrm{XLM}$-$\mathrm{R_{base}}$ & 30.4 & 22.4 & 10.6 & 13.5 & 19.2 \\
& & $\mathrm{MuRIL_{base}}$ & \textbf{34.0} & \textbf{22.7} & 10.8 & 14.7 & 20.6 \\
& & $\mathrm{XLM}$-$\mathrm{R_{large}}$ & 33.0 & 22.4 & \textbf{11.5} & 15.2 & 20.5 \\
& & $\mathrm{MuRIL_{large}}$ & 33.4 & 22.4 & 11.4 & 15.7 & \textbf{20.7} \\
\midrule
\multirow[b]{5}{*}{$\mathrm{M_{\tau = 1} \boldsymbol{\uparrow}}$} & \multirow{5}{*}{English} & IndicBERT & 39.2 & 44.2 & 36.6 & 28.4 & 37.1 \\
& & $\mathrm{XLM}$-$\mathrm{R_{base}}$ & 59.2 & 58.1 & 43.6 & 49.9 & 52.7  \\
& & $\mathrm{MuRIL_{base}}$ & 69.6 & 54.7 & 45.5 & 53.8 & 55.9 \\
& & $\mathrm{XLM}$-$\mathrm{R_{large}}$ & 61.2 & 60.3 & 46.6 & 56.6 &  56.2 \\
& & $\mathrm{MuRIL_{large}}$ & 68.2 & 58.6 & 47.4 & \textbf{57.9} & 58.0 \\
\cline{2-8} \multirow[t]{5}{*}{(Demographic)} & \multirow{5}{*}{Hindi} & IndicBERT & 61.0 & 61.6 & 39.8 & 29.9 & 48.1 \\
& & $\mathrm{XLM}$-$\mathrm{R_{base}}$ & 70.3 & 66.7 & 45.8 & 50.6 & 58.3 \\
& & $\mathrm{MuRIL_{base}}$ & \textbf{75.1} & \textbf{67.3} & 46.8 & 54.7 & 61.0 \\
& & $\mathrm{XLM}$-$\mathrm{R_{large}}$ & 74.4 & 66.8 & \textbf{49.4} & 53.2 & 60.9 \\
& & $\mathrm{MuRIL_{large}}$ & 74.8 & 66.5 & 49.2 & 54.6 & \textbf{61.3} \\
\midrule
\multirow{10}{*}{Gini Coeff. $\boldsymbol{\downarrow}$} & \multirow{5}{*}{English} & IndicBERT & 0.67 & 0.81 & 0.92 & 0.84 & 0.81 \\
& & $\mathrm{XLM}$-$\mathrm{R_{base}}$ & 0.61 & 0.76 & 0.88 & 0.83 & 0.77 \\
& & $\mathrm{MuRIL_{base}}$ & 0.59 & 0.76 & 0.88 & 0.83 & \textbf{0.76} \\
& & $\mathrm{XLM}$-$\mathrm{R_{large}}$ & 0.6 & 0.75 & 0.88 & 0.83 & 0.77 \\
& & $\mathrm{MuRIL_{large}}$ & 0.59 & 0.76 & 0.88 & 0.83 & 0.77 \\
\cline{2-8} & \multirow{5}{*}{Hindi} & IndicBERT & 0.68 & 0.8 & 0.91 & 0.83 & 0.81 \\
& & $\mathrm{XLM}$-$\mathrm{R_{base}}$ & 0.59 & 0.75 & 0.87 & 0.83 & \textbf{0.76} \\
& & $\mathrm{MuRIL_{base}}$ & 0.59 & 0.75 & 0.87 & 0.83 & \textbf{0.76} \\
& & $\mathrm{XLM}$-$\mathrm{R_{large}}$ & 0.59 & 0.76 & 0.88 & 0.83 & 0.77 \\
& & $\mathrm{MuRIL_{large}}$ & 0.59 & 0.75 & 0.87 & 0.83 & \textbf{0.76} \\
\midrule
\multirow{10}{*}{Efficiency $\boldsymbol{\uparrow}$} & \multirow{5}{*}{English} & IndicBERT & 53.6 & 50.2 & 56.7 & 66.0 & 56.6 \\
& & $\mathrm{XLM}$-$\mathrm{R_{base}}$ & 44.4 & 48.1 & 57.4 & 76.7 & 56.7 \\
& & $\mathrm{MuRIL_{base}}$ & 69.1 & 52.5 & 58.7 & \textbf{77.8} & 64.5 \\
& & $\mathrm{XLM}$-$\mathrm{R_{large}}$ & 44.4 & 48 & 57.3 & 76.3 & 56.5 \\ 
& & $\mathrm{MuRIL_{large}}$ & 63.1 & 51.8 & 57.8 & 75.7 & 62.1 \\
\cline{2-8} & \multirow{5}{*}{Hindi} & IndicBERT & 62.5 & 53.7 & 56.9 & 64.5 & 59.4 \\
& & $\mathrm{XLM}$-$\mathrm{R_{base}}$ & 48.3 & 50.0 & 58.5 & 75.9 & 58.2 \\
& & $\mathrm{MuRIL_{base}}$ & \textbf{69.8} & \textbf{56.2} & \textbf{59.8} & 77.3 & \textbf{65.8} \\
& & $\mathrm{XLM}$-$\mathrm{R_{large}}$ & 49.2 & 49.0 & 59.1 & 75.1 & 58.1 \\
& & $\mathrm{MuRIL_{large}}$ & 65.2 & 53.7 & 58.8 & 75.0 & 63.2 \\
\bottomrule
\end{tabular}}
\caption{\label{table:zero_shot}\emph{Zero-shot fine-tuning results.} Overall, $\mathrm{MuRIL_{large}}$ scores highest on the utility metrics, the Gini coefficient is relatively high across all models and both $\mathrm{MuRIL_{base}}$ and $\mathrm{MuRIL_{large}}$ are, on average, equal with regards to efficiency. Note that the metrics are computed considering all 23 languages as detailed in \refsec{sec:setup}. More discussions in \refsec{sec:zero_shot}.}
\end{table}

\end{document}